# When is an example a counterexample?

## [Extended Abstract]

Eric Pacuit
University of Maryland
TiLPS, Tilburg University
e.j.pacuit@uvt.nl

Arthur Paul Pedersen
Department of Philosophy
Carnegie Mellon University
ppederse@andrew.cmu.edu

Jan-Willem Romeijn
Faculty of Philosophy
Groningen University
j.w.romeijn@rug.nl

## ABSTRACT

In this extended abstract, we carefully examine a purported counterexample to a postulate of iterated belief revision. We suggest that the example is better seen as a failure to apply the theory of belief revision in sufficient detail. The main contribution is conceptual aiming at the literature on the philosophical foundations of the AGM theory of belief revision [1]. Our discussion is centered around the observation that it is often unclear whether a specific example is a "genuine" counterexample to an abstract theory or a misapplication of that theory to a concrete case.

## 1. INTRODUCTION

Starting with the seminal paper [1], the so-called AGM theory of belief revision has been extensively studied by logicians, computer scientists, and philosophers. The general setup is well-known, and we review it here to fix ideas and notation.

Let $K$ be a *belief set*, a set of propositional formulae closed under classical consequence representing an agent's initial collection of beliefs. Given a belief $\varphi$ that the agent has acquired, the set $K * \varphi$ represents the agent's collection of beliefs upon acquiring $\varphi$. A central project in the theory of belief revision is to study constraints on functions $*$ mapping a belief set $K$ and a propositional formula $\varphi$ to a new belief set $K * \varphi$. For reference, the key AGM postulates are listed in the Appendix (Section A). This simple framework has been analyzed, extended, and itself revised in various ways (see [2] for a survey of this literature), and much has been written about the status of its philosophical foundations (cf. [10, 21, 20]).

The basic AGM theory does not explicitly address the question of how to respond to a sequence of belief changes. The only salient constraint on iterated revision implied by the eight AGM postulates is the requirement that $(K * \varphi) * \psi \subseteq K * (\varphi \wedge \psi)$ provided $\neg \psi \notin K * \varphi$. [1] However, if $\neg \psi \in K * \varphi$, there is no constraint on $(K * \varphi) * \psi$. Various authors have attempted to rectify this situation, proposing additional rationality constraints on belief revision given a sequence of input beliefs [8, 9, 5, 15, 16, 18, 21, 6]. Two postulates which have been extensively discussed in the literature are the following constraints:

**I1** If $\psi \in \text{Cn}(\{\varphi\})$ then $(K * \psi) * \varphi = K * \varphi$

**I2** If $\neg \psi \in \text{Cn}(\{\varphi\})$ then $(K * \varphi) * \psi = K * \psi$

Each of these postulates have some intuitive appeal. Postulate **I1** demands if $\varphi \to \psi$ is a theorem (with respect to the background theory), then first learning $\psi$ followed by the more specific information $\varphi$ is equivalent to directly learning the more specific information $\varphi$. Postulate **I2** demands that first learning $\varphi$ followed by learning a piece of information $\psi$ incompatible with $\varphi$ is the same as simply learning $\psi$ outright. So, for example, first learning $\varphi$ and then $\neg \varphi$ should result in the same belief state as directly learning $\neg \varphi$. [2]

Many recent developments in this area have been offered on the basis of analyses of *concrete examples*. These range from toy examples—such as the infamous muddy children puzzle, the Monty Hall problem, and the Judy Benjamin problem—to everyday examples of social interaction. Different frameworks are then judged, in part, on how well they conform to the analyst's intuitions about the perceived relevant set of examples. This raises an important issue: Implicit assumptions about what the agents know and believe about the situation being modeled often guide the analyst's intuitions. In many cases, it is crucial to make these underlying assumptions explicit.

The following simple example illustrates the type of implicit assumption that we have in mind. There are two opaque boxes, labeled 1 and 2, each containing a coin. The believer is interested in the status of the coins in each box. Suppose that Ann is an expert on the status (heads up or tails up) of the coin in box 1 and that Bob is an expert on the status (heads up or tails up) of the coin in box 2. Currently the believer under consideration does not have an opinion about whether the coins are lying heads up or tails up in the boxes; more specifically, the believer thinks that all four possibilities are equally plausible. Suppose that both Ann and Bob report that their respective coins are lying tails up. Since both experts are trusted, this is what the believer believes. Now further suppose that there is a third expert, Charles, who is considered more reliable than both Ann and Bob. What should the believer think about the coin in box 2 after receiving a report from Charles that the coin in box 1 is lying heads up?

---

[1] By AGM 7 ($K * (\varphi \wedge \psi) \subseteq \text{Cn}(K * \varphi \cup \{\psi\})$) and AGM 8 ($\neg \psi \notin K * \varphi$ then $\text{Cn}(K * \varphi \cup \{\psi\}) \subseteq K * (\varphi \wedge \psi)$), we have $K * (\varphi \wedge \psi) = \text{Cn}(K * \varphi \cup \{\psi\})$ provided that $\neg \psi \notin K * \varphi$, whence by an application of AGM 3 (($K * \varphi) * \psi \subseteq \text{Cn}((K * \varphi) \cup \{\psi\})$), it follows that $(K * \varphi) * \psi \subseteq \text{Cn}(K * \varphi \cup \{\psi\})$ if $\neg \psi \notin K * \varphi$.

[2] Of course, one might object to this on the basis of the observation that if the believer is in a situation in which she is receiving inconsistent evidence, then she should recognize this and accordingly adopt beliefs about the source(s) of information. This issue of *higher order evidence* is interesting (cf. [7]), but we set it aside in this paper. We are interested in situations in which the believer never loses her trust in the process generating evidence. AGM theory may be the only theory applicable in such situations.

Of course, an answer to this question depends in part on the believer's initial opinions about the relationship between the coins in the two boxes. If the believer initially thinks that the status of the coins are independent and that the reports from Ann and Bob are independent, then she should believe that the coin in box 2 is lying tails up. However, if she has reason to think that the coins, or reports about the coins, are somehow correlated, upon learning that the coin in box 1 is lying heads up, she may be justified in changing her belief about the status of the coin in box 2.

Robert Stalnaker [21] has discussed the potential role that such *meta-information*, as illustrated in the above example, plays in the evaluation of proposed counterexamples to the AGM postulates. The general message is that once salient meta-information has been made explicit, many of the purported counterexamples to the AGM theory of belief revision do not demonstrate a failure of the theory itself, but rather a failure to *apply* the theory correctly and include all the relevant components in the model. [3] After an illuminating discussion of a number of well-known counterexamples to the AGM postulates, Stalnaker proposes two "genuine" counterexamples to postulates **I1** and **I2** for the theory of iterated belief revision. The conclusion Stalnaker draws in his discussion is that "... little of substance can be said about constraints on iterated belief revision at a level of abstraction that lacks the resources for explicit representation of meta-information" (pg. 189).

In this extended abstract, we carefully examine one of Stalnaker's purported counterexamples (Section 4), provide a model for it that complies with the AGM postulates, suggesting that it is again better seen as a failure to apply the theory of belief revision in sufficient detail. We end with a critical discussion of the opposition between genuine counterexamples and misapplications of the theory (Section 6).

## 2. STALNAKER'S EXAMPLE

An indicated in the introduction, Stalnaker [21] proposes counterexamples to both postulates **I1** and **I2**. In this extended abstract, we only have space to discuss one of the examples (the full paper has an extensive discussion of both examples). We discuss an example which is "clearer and a more decisive problem" for **I2**.

**Example**. Suppose that two fair coins are flipped and placed in two boxes. Two independent and reliable observers deliver reports about the status (heads up or tails up) of the coins in the opaque boxes. On the one hand, Alice reports that the coin in box 1 is lying heads up, and on the other hand, Bert reports that the coin in box 2 is lying heads up.

Two new independent witnesses, whose reliability trumps that of Alice's and Bert's, provide additional reports on the status of the coins. Carla reports that the coin in box 1 is lying tails up, and Dora reports that the coin in box 2 is lying tails up. Finally, Elmer, a third witness considered the most reliable overall, reports that the coin in box 1 is lying heads up.

Let $H_i$ be the proposition expressing the statement that the coin in box $i$ is lying heads up ($i = 1, 2$). Similarly, for $i = 1, 2$, let $T_i$ be the proposition expressing the statement that the coin in box $i$ is lying tails up. After the first belief revision, the belief set is $K' = K * (H_1 \wedge H_2)$, where $K$ is the agent's original set of beliefs. After receiving the reports, the belief set is $K' * (T_1 \wedge T_2) * H_1$. As Stalnaker suggests, since Elmer's report is irrelevant to the status of the coin in box 2, it seems natural to assume that $H_1 \wedge T_2 \in K' * (T_1 \wedge T_2) * H_1$.

Now to the hitch. Since $(T_1 \wedge T_2) \to \neg H_1$ is a theorem (given the background theory), by **I2** it follows that $K' * (T_1 \wedge T_2) * H_1 = K' * H_1$. Yet since $H_1 \wedge H_2 \in K'$ and $H_1$ is consistent with $H_2$, we must have $H_1 \wedge H_2 \in K' * H_1$, which yields a conflict with the assumption that $H_1 \wedge T_2 \in K' * (T_1 \wedge T_2) * H_1$.

Stalnaker diagnoses the situation as follows:

> ...[Postulate *I2*] directs us to take back the totality of any information that is overturned. Specifically, if we first receive information $\alpha$, and then receive information that conflicts with $\alpha$, we should return to the belief state we were previously in, before learning $\alpha$. But this directive is too strong. Even if the new information conflicts with the information just received, it need not necessarily cast doubt on *all* of that information.
> 
> (pg. 207–208)

It seems that, for lack of independent guidelines of how we must identify the component of the evidence that needs overturning to accommodate the new information, the epistemic advice provided by AGM conflicts with the intuitively correct answer.

But what are we to do with this apparent conflict? In what follows we attempt to model Stalnaker's puzzling example. This is a conceptual paper aiming to contribute to the literature on the philosophical foundations of the theory of belief revision (cf. [10, 21, 20]). Accordingly, it is not our main goal to extend this theory, resolve the problems, and be done with it. Our focus lies rather on the fact that it is unclear how to appropriately respond to a purported counterexample to a postulate of iterated belief revision. To illustrate, the foregoing example may be regarded as demonstrating either:

1. There is no suitable way to formalize the scenario in such a way that the AGM postulates (possibly including postulates of iterated belief revision) can be saved;

---

[3]This is not to say that there are no genuine conceptual problems with the AGM theory of belief revision. The point raised here is that it is often unclear what exactly a specific counterexample to an AGM postulate demonstrates about the abstract theory of belief revision. This is nicely explained by Stalnaker in his analysis of Hans Rott's well-known counterexample to various AGM postulates (see [20]):

> ... Rott seems to take the point about meta-information to explain why the example conflicts with the theoretical principles, whereas I want to conclude that it shows why the example does not conflict with the theoretical principles, since I take the relevance of the meta-information to show that the conditions for applying the principles in question are not met by the example.
> 
> (pg. 204)

2. The AGM framework can be made to agree with the scenario but does not furnish a systematic way to formalize the relevant meta-information; or

3. There is a suitable and systematic way to make the meta-information explicit, but this is something that the AGM framework cannot properly accommodate.

The first response is very drastic and, indeed, the models presented in the next section may be taken to show that the meta-information driving the belief change can be made suitably explicit. The second response to the example is already well-appreciated in the literature on belief revision (cf. the discussion of sources of evidence in [6] and "ontology" in [10]). Of interest to us for this paper is the third response, which is concerned with the absence of guidelines for *applying* the theory of belief revision.

In other words, we suggest that there is a problem with the AGM theory, and that this problem arises because a clear distinction between counterexample and misapplication has yet to be drawn. It is not clear from the theory of belief revision how its applications are supposed to be organized, and hence it is not clear whether the examples reveal shortcomings in the theory or rather in its application. Stalnaker suggests that his purported counterexamples turn on *independence*:

> There are different kinds of independence—conceptual, causal and epistemic—that interact, and one might be able to say more about constraints on rational belief revision if one had a model theory in which causal-counterfactual and epistemic information could both be represented. There are familiar problems, both technical and philosophical, that arise when one tries to make meta-information explicit, since it is self-locating (and auto-epistemic) information, and information about changing states of the world. (pg. 208)

Part of our response is to show how models from AGM belief revision can accommodate such considerations. In addition, we offer a different perspective on the third response to the counterexample.

This shift in perspective has a positive part and a negative part. On the one hand, we argue that probabilistic models can facilitate an explicit incorporation of meta-information underlying rational belief changes for many examples. Furthermore, these models can offer principled ways to distinguish genuine counterexamples from misapplications of the AGM theory of belief revision. In the example at hand, the salient categories are the event and report structure, and the belief states that range over them.

On the other hand, the connection between AGM theory and probabilistic models of belief revision offers an opportunity to exploit various insights from critical discussions concerning probabilistic models of belief dynamics. At the end of our paper, we critically discuss two such insights. The first insight draws attention to the absence of a genuine belief dynamics in probabilistic models: Bayesian models and their extensions are completely static. The second insight draw attention to a relationship with a result due to [13] identifying situations in which conditioning in so-called "naive" spaces matches conditioning in so-called "sophisticated" spaces (in which the relevant meta-information is made explicit).

## 3. A HEURISTIC TREATMENT

We begin with a heuristic treatment of Stalnaker's counterexample to postulate **I2**, serving to explain the role that the Bayesian model of Section 4 plays to respond to Stalnaker's challenge to an AGM theory of iterated belief revision.

The heuristic treatment is cast in the semantic model of AGM belief revision introduced in Grove's seminal paper [12]. The key idea is to describe the belief state of an agent as a set of possible worlds and a *plausibility relation* on this set of states (formally, a plausibility ordering is a reflexive, transitive and well-founded relation). To illustrate, in the example there are four possible worlds corresponding to the configurations of the coins in the two boxes. Initially, the believer considers all the configurations of the coins equally plausible. The agent *believes* any proposition implied by the set of most plausible worlds. A *belief revision policy* describes how to modify a plausibility ordering given a nonempty subset of the set of states (intuitively, this subset represents a belief that the agent has acquired).

A number of different belief revision policies have been identified and explored in the literature (cf. [20, 3, 22]). For our discussion of Stalnaker's counterexample, we focus on the so-called **radical upgrade** belief revision policy: If $\varphi$ is a set of worlds, the radical upgrade with $\varphi$, denoted $\Uparrow \varphi$, defines a new plausibility relation as follows: all the states in $\varphi$ become strictly more plausibility than all the states not in $\varphi$, while the ordering for states within $\varphi$ and outside of $\varphi$ remains the same.

Starting from an initial model in which the believer considers all positions of the coins equally plausible, the belief changes in Stalnaker's counterexample can be represented as follows:

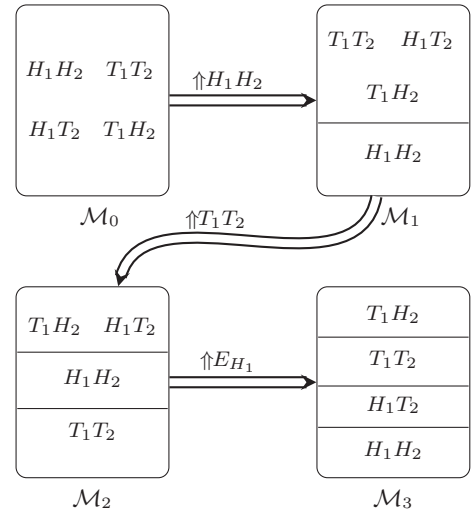

Interpret this diagram as follows: Each state is labeled by the position of the coins in the different boxs. The ordering is represented by the straight lines, with the states at the bottom the most plausible overall. For example, in model $\mathcal{M}_2$, since the state $T_1T_2$ is the most plausible overall, the agent believes that the coins in both boxs are lying tails up. Each transition corresponds to a radical upgrade with the identified set (we write $\Uparrow w$ instead of $\Uparrow \{w\}$, and the last transition is with the event $E_{H_1} = \{H_1H_2, H_1T_2\}$).

The above formalization highlights the crucial issue raised by Stalnaker's example: A side effect of first learning that both coins are lying heads up followed by learning that both coins are lying tails up is that the agent comes to believe that the coins in the two boxs are correlated. Note that in the third model $\mathcal{M}_2$, the state $H_1H_2$ is ranked more plausible than both $H_1T_2$ and $T_1H_2$. This is not necessarily problematic provided the agent's initial beliefs about the learning situation warrant such a conclusion. However, such meta-information is not made explicit in the description of the example. This leaves open the possibility of a counterintuitive reading of the example in which it is not rational for the believer to come to the conclusion that the coins are correlated.

Our suggestion is *not* that it is impossible to define a belief revision policy that incorporates the assumption that the believer takes it for granted that the coins are independent. Indeed, the following sequence represents such a belief revision policy:

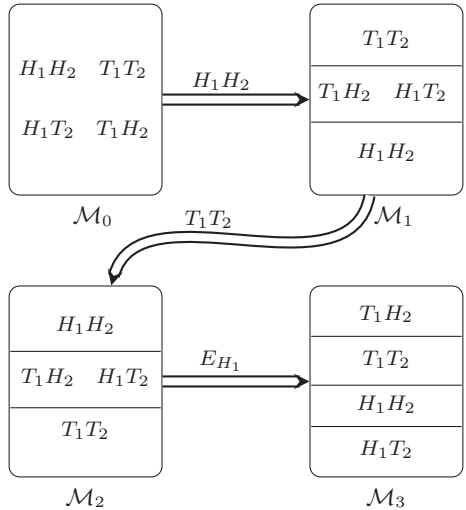

In the above formalization, each time the agent learns something about the position of the coins, the initial belief that the position of the coins are independent is retained. This leaves open the question of whether one can find a defensible belief revision policy generating such a sequence of belief changes. The models from sections 4 and 5 demonstrate that this question has an affirmative answer. The models provide a systematic way to explicitly describe the meta-information in the background underlying an application of the AGM theory of belief revision. However, as we argue in Section 6, this does not entirely resolve the issue that Stalnaker's raises.

## 4. BAYESIAN MODELS

In what follows, we sketch a Bayesian model formalizing salient meta-information in the example from Section 2. The model demonstrates that such information can be suitably captured in terms of a coherent set of revision rules. Of course, the model may be unsatisfactory to someone seeking to extend AGM belief revision theory with rules for iteration. Many Bayesian modeling choices, most notably concerning the representation of belief, are at odds with AGM theory. However, as we show in Section 5, the Bayesian model can be refined to cover belief revision policies in the style of AGM and, for example, Darwiche and Pearl [9] while retaining the formalization of salient intuitions. To be sure, we make no claim to a general model covering *all* cases and *all* potentially relevant meta-information. But in the example under discussion we think that insufficient detail has been offered to warrant any such claim in the first place.

**The Basic Formalization**

To fix ideas, a *Bayesian model* of Example 1 consists of an algebra over a set of states including all relevant propositions and a probability function expressing the agent's beliefs about these propositions as held at consecutive stages of her epistemic development. We lay down this basic structure, subsequently presenting three related probability functions accommodating meta-information.

The hypotheses at stake in the example concern the results of coin tosses in the two boxes, denoted $X_j^i$ with $i \in \{0,1\}$ for tails up 0 and heads up 1, respectively, and $j \in \{1, 2\}$ for boxes 1 and 2 respectively (here, for convenience, we use numerals rather than letters for the boxes). Furthermore, there are five reports, denoted $R_{jt}^i$, each with $i \in \{0, 1\}$ for a report of tails up or heads up and $j \in \{1, 2\}$ for boxes 1 and 2, and $t \in \{0, 1, 2, 3\}$ for the four update stages in the epistemic development of the agent. Letting $\mathcal{X}_j = \{X_j^0, X_j^1\}$ and $\mathcal{R}_{jt} = \{R_{jt}^0, R_{jt}^1\}$, we can write the state space $\Omega$ as

$$\Omega = \mathcal{X}_1 \times \mathcal{X}_2 \times (\prod_{t=1,2,3} \mathcal{R}_{1t} \times \mathcal{R}_{2t})$$

Thus a state $\omega \in \Omega$ is of the form

$$\omega = (X_1^{i_1}, X_2^{i_2}, R_{11}^{i_3}, R_{21}^{i_4}, R_{12}^{i_5}, R_{22}^{i_6}, R_{13}^{i_7}, R_{23}^{i_8}),$$

where $i_k \in \{0,1\}$ for each $k = 1, \ldots, 8$. We take the algebra $\mathcal{F}$ to be the power set of $\Omega$. Because reports and coins are mostly considered in pairs, we will use the abbreviations $X^{ik} = X_1^i \cap X_2^k$ and $R_t^{uv} = R_{1t}^u \cap R_{2t}^v$.

The beliefs of the agent are represented as probability functions over this algebra, $P_t : \mathcal{F} \to [0, 1]$. Summarizing the set of reports received up and until stage $t$ by the event $S_t$, and taking $X$ as the proposition of interest, the agent belief's are determined by Bayesian conditioning:

$$P_t(X) = P_0(X|S_t) = P_0(X)\frac{P_0(S_t|X)}{P_0(S_t)}.$$

In terms of the example, if we are interested in whether the coin in box 2 landed heads, $X_2^1$, the agent's belief state is a function of the probability conditional upon the reports of Alice and Bob, $P_1(X_2^1) = P_0(X_2^1|R_1^{11})$.

Since the two coins are fair and independent, the priors are $P_0(X^{ik}) = \frac{1}{4}$ for all $i, k = 0, 1$. We can now fill in the probability assignments to express the specific meta-information at stake in the example. The crucial point is that we can set the initial likelihoods in accordance with different intuitions about the meta-information in the example.

**The reports are independent**

In this case, after receiving the reports about the coins, the agent assigns high probability to the coin in box 1 lying heads up and the coin in box 2 lying tails up. According to the example, the the content of the reports are very probable, while the content of subsequent reports are even more probable, thereby cancelling out the impact of preceding reports. We can express this in the likelihoods of the hypotheses $X^{ik}$. For each combination of $j$ and $t$, let $Q$ be the event

$X_j^i \cap X_{3-j}^k \cap S_{t-1} \cap R_{(3-j)t}^v$. We have:

$$P_0(R_{jt}^u \mid Q) = \frac{1}{1+\gamma^t} \times \begin{cases} 1 & \text{if } u = i, \\ \gamma^t & \text{if } u \neq i. \end{cases} \quad (1)$$

The above expression fixes the probability of all report combinations given any state of the coins. Note that the likelihoods are independent of the reports $S_{t-1}$ of the preceding stage $t$, and that the reports $R_{jt}^u$ and $R_{(3-j)t}^v$ at stage $t$ are independent of each other too. Moreover, notice that $\gamma$ is the same for each report, expressing that reports at the same stage are equally reliable. Finally, the value $\gamma$ is close to zero, since the content of the reports are probable.[4]

These priors and likelihoods determine a full probability function $P_0$ over $\mathcal{F}$. By Bayes' rule, each probabilistic judgment at a later update stage is thereby fixed as well. We obtain the following posteriors:

| Time $t$ | 0 | 1 | 2 | 3 |
|---|---|---|---|---|
| After learning | $\top$ | $R_1^{11}$ | $R_2^{00}$ | $R_{13}^1$ |
| Odds for $X^{11}$ | 1 | 1 | $\gamma^2$ | $\gamma$ |
| Odds for $X^{10}$ | 1 | $\gamma$ | $\gamma$ | 1 |
| Odds for $X^{01}$ | 1 | $\gamma$ | $\gamma$ | $\gamma^3$ |
| Odds for $X^{00}$ | 1 | $\gamma^2$ | 1 | $\gamma^2$ |
| Prob. Evidence | 1 | $\frac{1}{4}$ | $\frac{1}{4}\gamma^2$ | $\frac{1}{4}\gamma^3$ |

After the first update stage with $R_1^{11}$, when the agent has received reports on the coins in both boxes, she is highly confident that both coins have landed heads. After the second pair of reports $R_2^{00}$, the agent is confident that both coins have landed tails. Finally, after Elmer's report $R_{13}^1$, the agent has revised her opinion about the coin in box 1 while leaving her opinion about the coin in box 2 unchanged.

Natural assumptions about the relationship amongst the reports, however, lead the agent to assign high probability to the event that both coins are lying heads up, as predicted by postulate **I2**.

**The reports are dependent**
The meta-information in the example may be such that Elmer's report also encourages the agent to change her mind about the coin in the second box. We can organize the Bayesian model in such a way that Elmer's report indeed has these consequences. Specifically, for $t < 3$ we may choose

$$P_0(R_t^{uv}|X^{ik} \cap S_{t-1}) =$$
$$\frac{1}{1+2\gamma^{1+t}+\gamma^{2+t}} \times \begin{cases} 1 & \text{if } u = i \text{ and } v = k, \\ \gamma^{1+t} & \text{if either } u \neq i, v = k \\ & \text{or } u = i, v \neq k, \\ \gamma^{2+t} & \text{if both } u \neq i, v \neq k. \end{cases}$$

and use the likelihood of Equation (1) for $t = 3$. This indicates that to some extent, the reports stand or fall together: if one of the reports at a particular stage is false, the other one is less reliable as well. In this case, deteriorating reliability is a factor $\gamma$, but we may organize the likelihoods differently to obtain different dependencies.

---

[4]In order for later reports to overrule earlier ones, it suffices to assume that for $t > 1$, the likelihoods are all $\frac{1}{1+\gamma^2}$. The present likelihoods indicate that reports also become increasingly reliable.

With the likelihood functions set up as above, we obtain the following posterior probability assignments:

| Time $t$ | 0 | 1 | 2 | 3 |
|---|---|---|---|---|
| After learning | $\top$ | $R_1^{11}$ | $R_2^{00}$ | $R_{13}^1$ |
| Odds for $X^{11}$ | 1 | 1 | $\gamma$ | 1 |
| Odds for $X^{10}$ | 1 | $\gamma^2$ | $\gamma^2$ | $\gamma$ |
| Odds for $X^{01}$ | 1 | $\gamma^2$ | $\gamma^2$ | $\gamma^4$ |
| Odds for $X^{00}$ | 1 | $\gamma^3$ | 1 | $\gamma^2$ |
| Prob. Evidence | 1 | $\frac{1}{4}$ | $\frac{1}{4}\gamma^3$ | $\frac{1}{4}\gamma^4$ |

In words, the first two belief changes of the agent are as before: for small $\gamma$ the beliefs shift from $X^{11}$ to $X^{00}$ with the pairs of reports. But after the final report about the coin in box 1, the agent also revises her opinion about the coin in box 2. Importantly, this arises not because the final report about the coin in box 1 has a direct bearing on our beliefs concerning the coin in box 2, but rather because in shifting the probability mass back towards $X_1^1$, the dominating factor in the probability for $X_2^1$ becomes $P_3(X_2^1|X_1^1)$. The belief dynamics is in this sense similar to the dynamics of so-called analogical predictions (cf. [19]).

It might be suggested that the foregoing analysis somehow fails to unfold what Stalnaker has in mind:

> Because my sources were independent, my belief revision policies, at one stage, will give priority to the [possibilities of one report being false] over the [possibility of both reports being false]. (Were I to learn that [one report] was wrong, I would continue to believe [the other report] and vice versa.). (p. 207)

In a footnote, Stalnaker adds, "nothing [in] the theory as it stands provides any constraints on what counts as a single input, or any resources for representing the independence of sources."

We agree with the assertion that the AGM theory does not itself furnish such resources and so in this sense the theory is lacking. Indeed, the assertion has likeminded friends, who when read air similar platitudes about other axiomatic theories offering minimal rationality principles, theories which also abstain from imposing substantial constraints on admissible states of belief. But the assertion does not also serve as a compelling excuse to advertise a poorly posed example as a counterexample. A good counterexample is packaged for self-assembly, equipped with details obviously relevant to its evaluation and relevant to its challenge in a meaningful debate about its significance.

In the present case, Stalnaker neglects to elaborate upon the form of independence relevant to the example, and he has not articulated the example in a way univocally suggesting a particular form of independence. Even for a familiar form of independence, the incomplete example may be supplemented with details which conform to a reading according to which the truth of either report is vastly more probable, independently of the truth of the other report. Yet the deficient example may also seek assistance with details which conform to another reading in which independence finds expression in terms of correlated reliability of the two reports, a reading consistent with independently varying reports.

Thus, a reply suggesting that our analysis somehow fails to unfold what Stalnaker has in mind simultaneously undertakes an obligation to articulate an argument supporting the claim that a relevantly different reading warrants recognition as an image of Stalnaker's thoughts—or at least recognition over our proposed readings.

**Elmer and Carla's reports are correlated**
With some imagination, we can also provide a model in which the pairs of reports are independent in the strictest sense, and in which Elmer's report is fully responsible for the belief change regarding both coins. To achieve this we employ the likelihoods of Equation (1) for the first two stages, but for the report of Elmer we use a rather gerrymandered set of likelihoods:

$$P_0(R_{13}^u | X^{ik} \cap R_2^{vw} \cap S_1) =$$
$$\frac{1}{1+\gamma^2+\gamma^3+\gamma^5} \times \begin{cases} 1 & \text{if } u = i \neq v \text{ and } w \neq k, \\ \gamma^2 & \text{if } u = i \neq v \text{ and } w = k, \\ \gamma^3 & \text{if } u \neq i = v \text{ and } w \neq k, \\ \gamma^5 & \text{if } u \neq i = v \text{ and } w = k. \end{cases}$$

Notice that the conditions on the right cover all combinations of indexes $i$, $k$, and $v$, but only half of their combinations with $u$. The likelihood for the opposite values of $u$ follow, because the probability of $R_{13}^u$ and $R_{13}^{1-u}$ must add up to 1.

Of course we may vary the exact conditions under which Elmer's report overturns the reports of both Carla and Dora. Moreover, as before, the specific numerical values chosen for the likelihoods only matter up to order of magnitude. The likelihoods given here make sure that until stage 2 the posteriors are as determined by Equation (1), and we have:

| Time $t$ | 2 | 3 |
|---|---|---|
| After learning | $R_1^{11} \wedge R_2^{00}$ | $R_{13}^1$ |
| Odds for $X^{11}$ | $\gamma^2$ | 1 |
| Odds for $X^{10}$ | $\gamma$ | $\gamma$ |
| Odds for $X^{01}$ | $\gamma$ | $\gamma^2$ |
| Odds for $X^{00}$ | 1 | $\gamma^3$ |
| Prob. Evidence | $\frac{1}{4}\gamma^2$ | $\frac{1}{4}\gamma^4$ |

Importantly, the likelihoods used to arrive at these posteriors square with the example provided by Stalnaker: Elmer's report is most probably reliable and indeed overturns the report by Carla. But the likelihoods are organized in such a way that they also overturn Dora's report under particular circumstances.

The full story of the agent might be that Carla and Dora use the same method to determine the state of their respective coins. Elmer almost always defers to Carla, unless he suspects something is amiss with her method, in which case he resorts to his own superior judgment. But he will only suspect something if in actual fact both Carla and Dora report falsely. Accordingly, conditional on both Carla's and Dora's reports being false, the agent expects Elmer's report to be true and hence at odds with Carla's. Similarly, on the condition that Carla's and Dora's report are both true, the agent considers it extremely probable that Elmer's report is true and in agreement with Carla's. Finally, if either Carla's or Dora's report is false, the agent considers Elmer's report to be most probably in line with Carla's, although less probably so if Carla's report is actually the false one. The agent imagines that Elmer tends to agree with Carla because he does not suspect anything is wrong with her method, and hence most likely defers to her.

Taking a step back, we admit that there will be many more ways of filling in the priors and likelihoods so as to represent particular aspects of the meta-information. However, the details of the full solution space need not concern us here. At this point, we simply note that the puzzle allows for Bayesian models that accommodate a range of intuitions.

## 5. NONSTANDARD PROBABILITY

As we have already noted, the Bayesian model in the previous section does not, by itself, offer a response to Stalnaker's challenge to the AGM-based theory of iterated belief revision. In this section, we explain precisely how the Bayesian model does in fact suggest a solution to Stalnaker's challenge which is in line with the standard assumption of the AGM theory of belief revision. The key step is to forge a connection between the AGM theory of belief revision and nonstandard probability measures. This connection between AGM and nonstandard probability measures is not surprising given the results in Appendix B of [17] relating non-monotonic logics with nonstandard probabilities.[5]

The key observation is that our discussion of the Bayesian model in the previous section and the conclusions we draw regarding Stalnaker's example do not depend on the specific values of the likelihoods used to calculate the agents' posterior beliefs. What is important are the order of magnitudes. Indeed, we can assume that the likelihoods are *arbitrarily small* and still derive the same qualitative consequences about belief change from the model. So if we represent the agents' full belief states by *nonstandard probability measures* and reinterpret the Bayesian model in those terms, we obtain a model that complies to the AGM postulates, and that nevertheless captures the role of meta-information in the desired way.

In the remainder of this section, we formally connect the nonstandard probability measures and the AGM theory of belief revision.

*Definition 1.* Let $\mathscr{A}$ be an algebra over a set of states $\Omega$, and let $^*\mathbb{R}$ be a nonstandard model of the reals. A $^*\mathbb{R}$-*valued probability function* on $\mathscr{A}$ is a mapping $\mu : \mathscr{A} \to {}^*\mathbb{R}$ satisfying the following properties:

(i) $\mu(A) \geq 0$ for every $A \in \mathscr{A}$;

(ii) $\mu(\Omega) = 1$;

(iii) For all disjoint $A, B \in \mathscr{A}$: $\mu(A \cup B) = \mu(A) + \mu(B)$.

We say that $\mu$ is *regular* if $\mu(A) > 0$ for every $A \in \mathscr{A}^\circ$ (where $\mathscr{A}^\circ$ is $\mathscr{A}$ without the emptyset).

---
[5]In what follows, we assume the reader is familiar with the basic concepts of nonstandard analysis. See [11] for a discussion.

For a limited hyperreal[6] $r \in {}^*\mathbb{R}$, let $\text{st}(r)$ be the unique real number infinitely close to $r$. Given a ${}^*\mathbb{R}$-valued probability function $\mu$ on an algebra $\mathscr{A}$, a collection $\mathscr{B} \subseteq \mathscr{A}$, an event $E \in \mathscr{A}$, and $r \in {}^*[0, 1]$, let:

$$\text{st}_r(\mu(\mathscr{B}|E)) := \begin{cases} \{A \in \mathscr{B} : \text{st}(\mu(A|E)) \geq r\} & \text{if } \mu(E) > 0; \\ \{A \in \mathscr{B} : \text{st}(\mu(A)) \geq r\} & \text{otherwise.} \end{cases}$$

When $\mathscr{A}$ is finite, we associate a set $K_\mu \in \mathscr{A}$ by setting:

$$K_\mu := \bigcap \text{st}_1(\mu(\mathscr{A}|\Omega)).$$

Observe that $K_\mu$ is consistent, since whenever $\text{st}(\mu(A)) = 1$ and $\text{st}(\mu(B)) = 1$ for some $A, B \in \mathscr{A}$, $\text{st}(\mu(A) + \mu(B) - \mu(A \cup B))) = \text{st}(\mu(A)) + \text{st}(\mu(B)) - \text{st}(\mu(A \cup B))) $ and so $\text{st}(\mu(A \cap B)) = 1$. Deifne an operator $*_\mu$ by setting for every $E \in \mathscr{A}$:

$$K_\mu *_\mu E := \begin{cases} \bigcap \text{st}_1(\mu(\mathscr{A}|E)) & \text{if } E \in \mathscr{A}^\circ; \\ \emptyset & \text{otherwise.} \end{cases}$$

As before, we omit subscripts when there is no danger of confusion. The precise connection between nonstandard probability measures and the AGM theory of belief revision is given by the following Proposition:

PROPOSITION 1. *Let $\mathscr{A}$ be a finite algebra over $\Omega$, and let $K \in \mathscr{A}^\circ$. Then $*$ is a belief revision operator for $K$ if and only if there is a regular ${}^*\mathbb{R}$-valued probability function $\mu$ on $\mathscr{A}$ such that $K = K_\mu$ and $* = *_\mu$.*

REMARK 1. *There is also an important connection with lexicographic probability systems in the sense of of [4] (cf. [14] for a full discussion). Given a finite algebra $\mathscr{A}$, there is an obvious one-to-one correspondence between conditional probability functions and lexicographic probability systems with disjoint supports. However, even on a finite algebra, there is no nontrivial one-to-one correspondence between lexicographic probability systems with disjoint supports and ${}^*\mathbb{R}$-valued probability functions. In addition, it is clear from the connection between lexicographic probability systems and conditional probability functions that in general while it may be that $*_P = *_{P'}$ and $K_P = K_{P'}$ it does not follow that $P = P'$ and indeed $*_\mu = *_{\mu'}$ and $K_\mu = K_\mu$ does not entail that $\mu = \mu'$.*

REMARK 2. *If one wishes to admit zero probabilities in the nonstandard setting, one may introduce the concept of a ${}^*\mathbb{R}$-valued (full) conditional probability function, thereupon defining a revision operator as for ${}^*\mathbb{R}$-valued probability functions, without the implicit requirement of regularity.*

With this connection between the AGM postulates and Bayesian models using nonstandard probability measures in place, let us return to the example of Stalnaker. In virtue of the connection, we can now reinterpret the Bayesian models of the example to obtain models for the dynamics of full belief that comply to the AGM postulates, while accommodating the role of meta-information in the right way. Specifically, we can make $\gamma$, the central parameter in the definition of the likelihoods in Section 4, arbitrarily small. In the tables detailing the probabilistic belief states from the example, we thereby set all entries to the extremal values 0 or 1. Depending on how the meta-information on the dependence of reports is spelled out, the belief dynamics thus retains the desired qualitative features.

---

[6]A hyperreal $r \in {}^*\mathbb{R}$ is said to be *limited* if there is a (standard) natural number n such that $|r| \leq n$.

## 6. COUNTEREXAMPLES VS MISAPPLICATIONS

The immediate upshot of the analysis in the previous section is that the puzzle from [21] does not present insurmountable problems for a theory of iterated belief revision. After all, the nonstandard probabilistic models present us with a formally worked out revision policy. In what follows we present an evaluation of what this analysis achieves and a more nuanced view on the status of the counterexample.

We would like to flag that it is not clear from his paper that Stalnaker thinks the example reveals fundamental limitations for the AGM theory of iterated belief revision. Therefore, rather than thinking that the models prove him wrong, we think of their potential virtue as being more positive: they indicate how a belief revision policy can incorporate particular kinds of meta-information. The nonstandard Bayesian models allow us to systematically accommodate information that, in the words of Stalnaker, pertains to the conceptual, causal, and epistemic relations among factual information items. We chose to focus on information that concerns the reliability of the reports provided, and the relations that obtain between those reliabilities. The claim is certainly not that we have thereby exhausted the meta-information that may be relevant in the puzzle. But at least we have illustrated how such meta-information may come into play in a Bayesian model complying to the basic AGM postulates.

Our illustration allows us to draw some general lessons about the balance between counterexamples and misapplications in the context of modeling belief dynamics. The models bring out how tentative counterexamples can be overcome by carefully explicating various aspects from the problematic example case. Several categories of analysis deserve further analysis. First, a proper conceptualization of the event and report structure is crucial: we need a sufficiently rich structure of events, messages, epistemic states and the like to express all the meta-information. Note, however, that such a conceptualization is never part and parcel of the theory about the belief dynamics itself. A theory needs to be able to accommodate the conceptualization, but other than that it hardly counts in favor of a theory that the modeler gets this conceptualization right. Secondly it stands out that we must allow ourselves all the requisite tools for representing beliefs. In the puzzles at hand, the language must allow us to separate reports by different agents from the content of the reports. And most importantly, the expressions of belief must allow for some notion of graded disbelief or, as one may also put it, memory.

It may be thought that we think any purported counterexample can in the end be accommodated by a nonstandard Bayesian model or similar structure, and that any type of meta-information is amenable to the kind of treatment just illustrated. Are there any genuine counterexamples to be had, or do we want to reduce everything to misapplication? Here we get to the negative part of our perspective on the discussion on counterexamples to the theory of belief revision. We do believe that the theory of AGM belief revision and its probabilistic counterpart may have fundamental lim-itations. In the remainder of this section, we first consider one specific aspect in which the probabilistic models we have provided miss the mark, suggesting that the counterexamples still stand unresolved. Secondly, we sketch some results from [13] on how far Bayesian models can come in capturing meta-information, and thereby provide a prospect for the construction of counterexamples.

Researchers coming from the literature on iterated belief revision and current dynamic logics of belief revision, may be unsatisfied with the Bayesian models presented here as a solution to Stalnaker's counterexample. The Bayesian models represent belief change by conditioning (or one of its generalization, such as Jeffrey or Adam's conditioning). It can be argued that this is not a truly *dynamic* model of belief change. If the challenge from Stalnaker's example is to capture the belief changes while maintaining the dynamic character, then the Bayesian models presented here do not present a proper rebuttal. In turn, we suggest that purported counterexample must place more emphasis on the dynamic aspect of the problem.

This raises an interesting question for future research. There seems to be a trade-off between a rich set of states and event structure, and a rich theory of "doxastic actions" (eg., as found in the literature on dynamic logics of belief revision [22, 3]). How should we resolve this trade-off when analyzing counterexamples to postulates that are intended to apply to belief changes over time. More generally, what is it about a dynamic model of belief revision that makes it truly dynamic?

We now turn to another prospect of genuine counterexamples to the theory of belief revision. For readers familiar with the flexibility of Bayesian models, it is not surprising that they allow us to formalize the relevant aspects of the meta-information in Stalnaker's example. The challenge seems rather to find out under what conditions we can ignore the meta-information, which is often not specified in the description of an example. Halpern and Grünwald [13] identify such a condition for Bayesian models, called *coarsening at random* or CAR for short. They study situations in which conditioning on a "naive" space gives the same results as conditioning on a "sophisticated" space'. Generally speaking, a "sophisticated" space is one that includes an explicit description of the relevant meta-information (eg., the reports from the sources and how they may be correlated). In the full paper we show how to apply this condition to the AGM framework. Or more precisely, we generalize the condition from [13] to nonstandard probability measures and then use the general link between AGM and nonstandard probability measures to apply the condition to AGM. In this extended abstract, however, we only have space to sketch the main idea of our result.

As indicated, CAR tells us how probabilities in naive and sophisticated spaces need to relate in order for updating by conditionalisation to be a correct inference rule in the naive space. But recall that in the generalization to nonstandard probability models, such updates follow the AGM postulates. The direct link to the examples given above is that whenever we run into a tentative counterexample, we can blame the failures of the update rule on a failure of CAR and start the repairs by building a more sophisticated state space. In cases like that, the culprit is arguably the application of the theory of belief revision: in a more refined space the update will again comply to AGM.

The same line of reasoning can now be used to clarify when misapplication turns into counterexample. In particular, we might argue that genuine counterexamples to AGM are cases in which we cannot blame failures of CAR. We see at least two ways in which this might happen. First, we might simply have no formalization of the problem case that allows for a representation of the update as a conditioning operation. Perhaps we cannot construct a sophisticated space, because the report or event structure does not allow for the definition of a partition of possible learning events. And second, it may so happen that AGM outputs an unintuitive epistemic state, even though we have employed an independently motivated formalization of the problem case. Attempts to redo the construction of a sophisticated space, just in order to remedy the failure of CAR, will be contrived. Instead, it may seem fair to blame the theory of belief revision itself.

In sum, we submit that the condition CAR may help us to formulate a principled distinction between misapplications of, and genuine counterexamples against a theory for belief dynamics. The appropriate response to the former is to refine the model and run the belief dynamics on the more refined space. Genuine counterexamples of the theory, on the other hand, are such that refinements of the model are impossible or contrived.

# 7. CONCLUSION

Our contribution in this paper is conceptual. First we have made explicit the meta-information implicit in one of Stalnaker's counterexample to a postulate of iterated belief revision. We have done so by identifying the salient meta-information in a heuristic model using plausibility orderings, by formalizing this information in a Bayesian model, and finally by generalizing this Bayesian model towards nonstandard probability models and showing that such models comply to the AGM postulates. This link between AGM and nonstandard probabilities allows us to use the characterization of the CAR condition to classify when a more refined state space can be used to explain the counterexample. Genuine counterexamples to AGM and iterated belief revision are cases when we cannot blame the structure of the state space. Our eventual goal is to develop a framework in which this intuition can be used to classify purported counterexamples.

# 8. REFERENCES

[1] C. E. Alchourrón, P. Gärdenfors, and D. Makinson. On the logic of theory change: Partial meet contraction and revision functions. *Journal of Symbolic Logic*, 50:510 – 530, 1985.

[2] H. Arló-Costa and A. P. Pedersen. Belief revision. In L. Horsten and R. Pettigrew, editors, *Continuum Companion to Philosophical Logic*. Continuum Press, 2011.

[3] A. Baltag and S. Smets. Conditional doxastic models: A qualitative approach to dynamic belief revision. In G. Mints and R. de Queiroz, editors, *Proceedings of WOLLIC 2006, Electronic Notes in Theoretical Computer Science*, volume 165, pages 5 – 21, 2006.

[4] L. Blume, A. Brandenburger, and E. Dekel. Lexicographic probabilities and choice under uncertainty. *Econometrica*, 59(1):61 – 79, 1991.

[5] C. Boutilier. Iterated revision and minimal revision of conditional beliefs. *Journal of Philosophical Logic*, 25:262 – 304, 1996.

[6] S. Chopra, A. Ghose, T. Meyer, and K.-S. Wong. Iterated belief change and the recovery axiom. *Journal of Philosophical Logic*, 37:501– 520, 2008.

[7] D. Christensen. Higher-order evidence. *Philosophy and Phenomenological Research*, 81(1):185 – 215, 2010.

[8] A. Darwiche and J. Pearl. On the logic of iterated belief revision. In *Proceedings of the 5th conference on Theoretical aspects of reasoning about knowledge*, pages 5–23, 1994.

[9] A. Darwiche and J. Pearl. On the logic of iterated belief revision. *Artificial Intelligence*, 89(1–2):1–29, 1997.

[10] N. Friedman and J. Y. Halpern. Belief revision: A critique. In L. C. Aiello, J. Doyle, and S. Shapiro, editors, *Proceedings of the Fifth International Conference on Principles of Knowledge Representation and Reasoning, KR'96*, pages 421–431. Morgan Kaufmann, Cambridge, Mass., November 1996.

[11] R. Goldblatt. *Lectures on Hyperreals: An Introduction to Nonstandard Analysis*. Springer, 1998.

[12] A. Grove. Two modellings for theory change. *Journal of Philosophical Logic*, 17(2):157–170, 1988.

[13] P. Grünwald and J. Y. Halpern. Updating probabilities. *Journal of Artificial Intelligence Research*, 19:243 – 278, 2003.

[14] J. Y. Halpern. Lexicographic probability, conditional probability, and nonstandard probability. *Games and Economic Behavior*, 68(1):155 – 179, 2010.

[15] H. Katsuno and A. O. Mendelzon. Propositional knowledge base revision and minimal change. *Artifcial Intelligence*, 52:263 – 294, 1991.

[16] D. Lehman. Belief revision, revised. In *Fourteenth International Joint Conference on Artificial Intelligence*, pages 1534–1541, 1995.

[17] D. Lehman and M. Magidor. What does a conditional knowledge base entail? *Artificial Intelligence*, 55(1):1 – 60, 1992.

[18] A. Nayak, M. Pagnucco, and P. Peppas. Dynamic belief revision operators. *Artificial Intelligence*, 146:193 – 228, 2003.

[19] J.-W. Romeijn. Analogical predictions for explicit similarity. *Erkenntis*, 64:253 – 280, 2006.

[20] H. Rott. *Change, Choice and Inference: A Study of Belief Revision and Nonmonotonic Reasoning*. Oxford University Press, 2001.

[21] R. Stalnaker. Iterated belief revision. *Erkentnis*, 70:189 – 209, 2009.

[22] J. van Benthem. Dynamic logic for belief revision. *Journal of Applied Non-classical Logics*, 17(2):129–155, 2007.

# APPENDIX
# A. THE AGM POSTULATES

In what follows, $K$ is a deductively closed and consistent set of propositional formulas and $\varphi, \psi$ are propositional formulas. Furthermore, $\text{Cn}(X)$ denotes the propositional consequences of a set $X$ of formulas.

The following are the *basic* revision postulates of AGM belief revision:

| | | |
|---|---|---|
| AGM 1 | (Closure) | $K * \varphi = \text{Cn}(K * \varphi)$. |
| AGM 2 | (Success) | $\varphi \in K * \varphi$. |
| AGM 3 | (Inclusion) | $K * \varphi \subseteq \text{Cn}(K \cup \{\varphi\})$. |
| AGM 4 | (Vacuity) | If $\neg\varphi \notin K$, then $\text{Cn}(K \cup \{\varphi\}) \subseteq K * \varphi$. |
| AGM 5 | (Consistency) | If $\text{Cn}(\{\varphi\}) \neq \text{For}(\mathcal{L})$, then $K * \varphi \neq \text{For}(\mathcal{L})$. |
| AGM 6 | (Extensionality) | If $\text{Cn}(\{\varphi\}) = \text{Cn}(\{\psi\})$, then $K * \varphi = K * \psi$. |

These six basic postulates are elementary requirements of belief revision and taken by themselves are much too permissive. Additional postulates are required to rein in this permissiveness and to reflect a conception of *relational* belief revision.

| | |
|---|---|
| AGM 7 | $K * (\varphi \wedge \psi) \subseteq \text{Cn}((K * \varphi) \cup \{\psi\})$. |
| AGM 8 | If $\neg\psi \notin K * \varphi$, then $\text{Cn}(K * \varphi \cup \{\psi\}) \subseteq K * (\varphi \wedge \psi)$. |

In the context of a propositional model (where $K$ is now a set of states and $E, F$ are also sets of states), all eight postulates may be reduced to four:

| | |
|---|---|
| Success ($*1$) | $K * E \subseteq E$. |
| Conditionalization ($*2$) | If $K \cap E \neq \emptyset$, then $K * E = K \cap E$. |
| Consistency ($*3$) | If $E \neq \emptyset$, then $K * E \neq \emptyset$. |
| (Arrow) ($*4$) | If $(K * E) \cap F \neq \emptyset$, then $(K * E) \cap F = K * (E \cap F)$. |

We say that $*$ is a *belief revision operator* for $K$ if it satisfies postulates $(*1) - (*4)$. See [20] for an extended discussion.

## A.1 AGM and conditional probability

In order to facilitate the relationship between the AGM theory of belief revision and nonstandard probability measures, we point out the relationship between AGM and *conditional probability measures*.

*Definition 2.* Let $\mathscr{A}$ be an algebra over a set of states $\Omega$. A (*full*) *conditional probability function* on $\mathscr{A}$ is a mapping $P : \mathscr{A} \times \mathscr{A} \to \mathbb{R}$ satisfying the following properties:

(i) $P(\cdot|E)$ is a finitely additive probability function for every $E \in \mathscr{A}^\circ$;

(ii) $P(A|E) = 1$ for every $A, E \in \mathscr{A}$ such that $E \subseteq A$;

(iii) For all $A, B, E \in \mathscr{A}$ such that $A \subseteq B \subseteq E$:
$$P(A|E) \;=\; P(A|B)P(B|E).$$

Here $\mathscr{A}^\circ$ is $\mathscr{A}$ without the null event $\emptyset$. Observe that $P(\cdot|\emptyset) \equiv \mathbf{1}$.

Given a conditional probability function $P$ on a finite algebra $\mathscr{A}$, we associate a set $K_P \in \mathscr{A}$ by setting:

$$K_P \;:=\; \text{supp } P(\cdot|\Omega),$$

where as usual supp $P(\cdot|E)$ denotes the probabilistic support of $P(\cdot|E)$, i.e., the smallest set in $\mathscr{A}$ receiving probability one on the condition that $E$ obtains. Define a belief revision operator $*_P$ by setting for every $E \in \mathscr{A}$:

$$K_P *_P E \; := \; \text{supp } P(\cdot|E).$$

We drop subscripts when the context is clear. The following is easily verified.

LEMMA 1. *Let $P$ be a conditional probability function on a finite algebra $\mathscr{A}$ over $\Omega$. Then $*_P$ is a belief revision operator for $K_P$.*

We also have the converse for consistent $K$, resulting in the following proposition.

PROPOSITION 2. *Let $\mathscr{A}$ be a finite algebra over $\Omega$, and let $K \in \mathscr{A}^\circ$. Then $*$ is a belief revision operator for $K$ if and only if there is a conditional probability function $P$ on $\mathscr{A}$ such that $K = K_P$ and $* = *_P$.*

This concludes our exposition of AGM in relation to conditional probability functions.

## A.2 Nonstandard probability

We now show how a similar link can be forged between AGM and nonstandard probability functions, admitting the possibility of arbitrarily small probability values.

Recall that given a $^*\mathbb{R}$-valued probability function $\mu$ on an algebra $\mathscr{A}$, a collection $\mathscr{B} \subseteq \mathscr{A}$, an event $E \in \mathscr{A}$, and $r \in {}^*[0, 1]$:

$$\text{st}_r(\mu(\mathscr{B}|E)) \; := \; \begin{cases} \{A \in \mathscr{B} : \text{st}(\mu(A|E)) \geq r\} & \text{if } \mu(E) > 0; \\ \{A \in \mathscr{B} : \text{st}(\mu(A)) \geq r\} & \text{otherwise}. \end{cases}$$

Where $\mathscr{A}$ is finite, we associate a set $K_\mu \in \mathscr{A}$:

$$K_\mu \; := \; \bigcap \text{st}_1(\mu(\mathscr{A}|\Omega)).$$

Define an operator $*_\mu$ by setting for every $E \in \mathscr{A}$:

$$K_\mu *_\mu E \; := \; \begin{cases} \bigcap \text{st}_1(\mu(\mathscr{A}|E)) & \text{if } E \in \mathscr{A}^\circ; \\ \emptyset & \text{otherwise}. \end{cases}$$

As before, we omit subscripts when there is no danger of confusion.

LEMMA 2. *Let $\mu$ be a regular $^*\mathbb{R}$-valued probability function on a finite algebra $\mathscr{A}$ over $\Omega$. Then $*_\mu$ is a belief revision operator for $K_\mu$.*

PROOF. Clearly postulates $(*1)$ and $(*3)$ are satisfied. While routine, for the sake of completeness we verify postulates $(*2)$ and $(*4)$ in turn.

($*2$) Suppose that $K \cap E \neq \emptyset$. Let $A \in \mathscr{A}$ be such that $K * E \subseteq A$. Then $\text{st}(\mu(A|E)) = 1$. Observe that:

$$\begin{aligned}
\text{st}(\mu(A \cup E^c)) &= \text{st}(\mu(E)\mu(A|E) + \mu(E^c)) \\
&= \text{st}(\mu(E))\text{st}(\mu(A|E)) + \text{st}(\mu(E^c)) \\
&= \text{st}(\mu(E) + \mu(E^c)) \\
&= 1.
\end{aligned}$$

It follows that $K \cap E \subseteq A$, establishing that $K \cap E \subseteq K * E$. Now let $A \in \mathscr{A}$ be such that $K \cap E \subseteq A$. Then $K \subseteq A \cup E^c$ and so $\text{st}(\mu(A \cup E^c)) = 1$, whence:

$$\begin{aligned}
\text{st}(\mu(E)) &= \text{st}(\mu(A^c \cap E)) + \text{st}(\mu(A \cap E)) \\
&= \text{st}(\mu(A \cap E)) \\
&= \text{st}(\mu(A|E))\text{st}(\mu(E)).
\end{aligned}$$

Thus, since $K \cap E \neq \emptyset$, it follows that $\text{st}(\mu(E)) > 0$ and therefore $\text{st}(\mu(A|E)) = 1$, whereby $K * E \subseteq A$. Hence, $K * E \subseteq K \cap E$.

($*4$) Suppose that $(K*E) \cap F \neq \emptyset$. Let $A \in \mathscr{A}$ be such that $K * (E \cap F) \subseteq A$. Then $\text{st}(\mu(A|E \cap F)) = 1$, so:

$$\begin{aligned}
\text{st}(\mu(A \cup F^c|E)) &= 1 - \text{st}(\mu((A^c \cap F)|E)) \\
&= 1 - \text{st}(\mu(A^c|F \cap E))\text{st}(\mu(F|E)) \\
&= 1.
\end{aligned}$$

Hence, $(K*E) \cap F \subseteq A$, showing that $(K*E) \cap F \subseteq K*(E \cap F)$. Now let $A \in \mathscr{A}$ be such that $(K*E) \cap F \subseteq A$. Then $\text{st}(\mu(A \cup F^c|E)) = 1$, and since $(K * E) \cap F \neq \emptyset$, it follows that $\text{st}(\mu(F|E)) \neq 0$, so:

$$\begin{aligned}
\text{st}(\mu(A|E \cap F))) &= 1 - \text{st}(\mu(A^c|E \cap F))) \\
&= 1 - \text{st}(\frac{\mu((A^c \cap F)|E)}{\mu(F|E)}) \\
&= 1 - \frac{\text{st}(\mu((A^c \cap F)|E))}{\text{st}(\mu(F|E))} \\
&= 1.
\end{aligned}$$

Therefore, $K*(E \cap F) \subseteq A$, so $K*(E \cap F) \subseteq (K*E) \cap F$, as desired. $\square$

PROPOSITION 3. *Let $\mathscr{A}$ be a finite algebra over $\Omega$, and let $K \in \mathscr{A}^\circ$. Then $*$ is a belief revision operator for $K$ if and only if there is a regular $^*\mathbb{R}$-valued probability function $\mu$ on $\mathscr{A}$ such that $K = K_\mu$ and $* = *_\mu$.*

PROOF. The 'if' part has been established in Lemma 2. We turn to the 'only if' part. Let $P$ be a conditional probability function on $\mathscr{A}$ such that $K = K_P$ and $* = *_P$, as given by Proposition 2. Then there is a partition $(\pi_m)_{m<n}$ of $\Omega$ in $\mathscr{A}$ and a sequence $(\mu_m)_{m<n}$ of real-valued probability functions on $\mathscr{A}$ such that:

(a) $\pi_m = \text{supp } \mu_m$ for each $m < n$;
(b) $P(\cdot|E) = \mu_{\min\{m:E\cap \pi_m \neq \emptyset\}}(\cdot|E)$ for every $E \in \mathscr{A}^\circ$.

Define a regular $^*\mathbb{R}$-valued probability function $\mu$ on $\mathscr{A}$ by setting for every $A \in \mathscr{A}$:

$$\mu(A) \; := \; \mu_0(A) + \sum_{0<m<n} (\mu_m(A) - \mu_0(A))\epsilon^m,$$

where $\epsilon$ is a positive infinitesimal. We claim that (i) $K_P = K_\mu$ and that (ii) $*_P = *_\mu$. Clearly $K *_P \emptyset = K *_\mu \emptyset$. Now let $E \in \mathscr{A}^\circ$. Set $m_0 := \min\{m : E \cap \pi_m \neq \emptyset\}$, and for each $m < n$, let $\nu_m := \mathbf{0}$ if $m = 0$ and $\mu_m$ otherwise. Then for every $A \in \mathscr{A}$:

$$\begin{aligned}
&\text{st}(\mu(A|E)) \\
&= \text{st}(\frac{\mu_0(A \cap E) + \sum_{0<m<n}(\mu_m(A \cap E) - \mu_0(A \cap E))\epsilon^m}{\mu_0(E) + \sum_{0<m<n}(\mu_m(E) - \mu_0(E))\epsilon^m}) \\
&= \text{st}(\frac{\mu_{m_0}(A|E) + \sum_{m_0<m<n} \frac{\mu_m(A \cap E) - \nu_m(A \cap E)}{\mu_{m_0}(E)}\epsilon^{m-m_0}}{1 + \sum_{m_0<m<n} \frac{\mu_m(E) - \nu_m(E)}{\mu_{m_0}(E)}\epsilon^{m-m_0}}) \\
&= \frac{\text{st}(\mu_{m_0}(A|E) + \sum_{m_0<m<n} \frac{\mu_m(A \cap E) - \nu_m(A \cap E)}{\mu_{m_0}(E)}\epsilon^{m-m_0})}{\text{st}(1 + \sum_{m_0<m<n} \frac{\mu_m(E) - \nu_m(E)}{\mu_{m_0}(E)}\epsilon^{m-m_0})} \\
&= \mu_{m_0}(A|E).
\end{aligned}$$

Then by property (b), claims (i) and (ii) follow, thereby establishing the desired conclusion. $\square$

REMARK 3. *The sequence $(\mu_m)_{m<n}$ of real-valued probability functions in the proof of Proposition 3 is a lexicographic probability system as discussed in Remark 1.*